\documentclass{article}

\usepackage[utf8]{inputenc}
\usepackage{amsmath, amsfonts, dsfont}
\usepackage{graphicx}

\usepackage{longtable, booktabs}

\usepackage[numbers, square]{natbib}

\usepackage{geometry}
\begin{document}

\title{Efficient and robust calibration of the Heston option pricing model for American options using an improved Cuckoo Search Algorithm}

\author{Stefan Haring \and Ronald Hochreiter}

\maketitle

\begin{abstract}
In this paper an improved Cuckoo Search Algorithm is developed to allow for an efficient and robust calibration of the Heston option pricing model for American options. Calibration of stochastic volatility models like the Heston is significantly harder than classical option pricing models as more parameters have to be estimated. The difficult task of calibrating one of these models to American Put options data is the main objective of this paper. Numerical results are shown to substantiate the suitability of the chosen method to tackle this problem.
\end{abstract}

\section{Introduction}
\label{sec:introduction}

The classical textbook example for pricing options is the famous Black-Scholes model, see \cite{black1973pricing}. Since its creation in 1973, it has caused a dramatic increase in options trading because of its relatively simple usability. It provides closed-form solutions for European Call and Put options as well as American Call options. For American Put options, no closed-form solution exists because an optimal stopping time problem has to be solved. 

Over the years, criticism of the Black-Scholes model has arisen since it does not accurately capture the behaviour of option prices in the market. For example, the well-known volatility smile, which shows that deep in the money and out the money options have higher prices (and therefore higher implied volatility), is not taken into account in the Black-Scholes model since one of its main assumptions is constant volatility.

To accurately reflect market behaviour, option pricing models in which volatility changes over time, so-called stochastic volatility models, have been created. In these models, the volatility is itself following a process that changes with time $t$.  Some examples for these models are the Heston model \cite{Heston}, the CEV model \cite{CEV1} \cite{CEV2} and the Chen model \cite{Chen}.

Calibration of these stochastic volatility models is significantly harder than for the Black-Scholes model, since more parameters have to be estimated. The difficult task of calibrating one of these models to American Put options data is the main objective of this paper. We will focus on the Heston model as it is widely known and cited. In his seminal paper, Heston specified his stochastic volatility model for option prices as following:

\begin{equation}
\frac{dS(t)}{S(t)} = \mu dt + \sqrt{V(t)}dW_1
\end{equation}

\begin{equation}
dV(t) = \kappa(\theta - V(t))dt + \sigma \sqrt{V(t)}dW_2,
\end{equation}

where $W_1$ and $W_2$ are Wiener processes that are correlated by $dW_1 * dW_2 = \rho dt$. The parameters used for obtaining the variance process are the long run variance $\theta$, the mean reversion rate $\kappa$ (the rate at which the volatility reverts to the long run variance) and the volatility of variance, $\sigma$.

Pricing options under the Heston model is not straight-forward when it comes to American option prices. The price of an American Call option is the same as for a European Call (because an early exercise is never optimal), however for American Put options no closed-form solution exists and therefore one has to use numerical approximations. There are many different methods in the literature for the approximations of these options such as the GARCH lattice approach as proposed by Ritchken and Trevor \cite{Garch}, tree-models presented by Beliaeva and Nawalkha \cite{Beliaeva} or Monte Carlo Simulation developed by Longstaff and Schwarz \cite{Longstaff}.

What is more important than choosing one of the many possible ways of how to price American options under the Heston model is an efficient and robust way to calibrate the model to actual data. After all, using the model in practice for means of forecasting and pricing is the main objective. The goal of this paper is to fill the gap existing in research carried out on the Heston model. At the time of writing, no rigorously tested method of calibrating the model for American Call options has been published.

The main reason is that while pricing options under this model is already a difficult task in itself, calibrating the model to a set of given option prices is even more challenging. In the standard model (keeping the interest rate fixed), there are five parameters that have to be calibrated: the aforementioned $\kappa$, $\theta$ and $\sigma$ as well as the starting value for the variance process $V_0$ and the correlation between the stock price and the variance, $\rho$.

What has been done, however, is calibrating the Heston model for European and Asian options. Gilli and Schumann show the effectiveness in using Heuristics and Evolutionary Algorithms when calibrating the Heston Model for European options \cite{Gilli1}, \cite{Gilli2}. Collier gives an overview of algorithms that can be used for optimising a noisy objective function, in this case an Monte Carlo simulation, to calibrate the Heston model to Asian options data \cite{Collier}. 

In this paper a naturally-inspired algorithm, namely Cuckoo Search, will be used to tackle the calibration problem. Naturally-inspired algorithms (also known as Evolutionary Algorithms) have been used on a lot of difficult optimization problems in Finance and have shown great promise (see for example \cite{Natural1}, \cite{Natural2}, \cite{Natural3}, \cite{Natural4}). See also relevant examples published in this journal including \cite{dash2014self}, \cite{dye2012finite}, \cite{mishra2014comparative}, \cite{beyer2014evolution}, and \cite{pai2014metaheuristic}.

The paper is structured as follows. Section \ref{sec:problem} gives a detailed overview over the problem and section \ref{sec:cuckoo} will provide a brief introduction to the Cuckoo Search Algorithm. In section \ref{sec:implementation}, the chosen implementation for the algorithm will be explained while in Section \ref{sec:results} numerical results will be presented. Section \ref{sec:conclusion} concludes this paper.

\section{Calibrating pricing models}
\label{sec:problem}

When calibrating a pricing model, the task is to find the parameters that minimize the difference between the model price and the actual observed price. This is reflected in the following optimization problem:

\begin{equation}
\min \sum_{i=1}^{N} (P_{i}^{model} - P_{i}^{obs})^2,
\label{eq:objfunc}
\end{equation}

where $N$ is the total number of observed prices. The optimization problem can also be specified differently (\cite{Natural4}), however the approach given in Equation (\ref{eq:objfunc}) is used as the objective function throughout this paper.

\begin{figure}
\centering
\includegraphics[width=80mm]{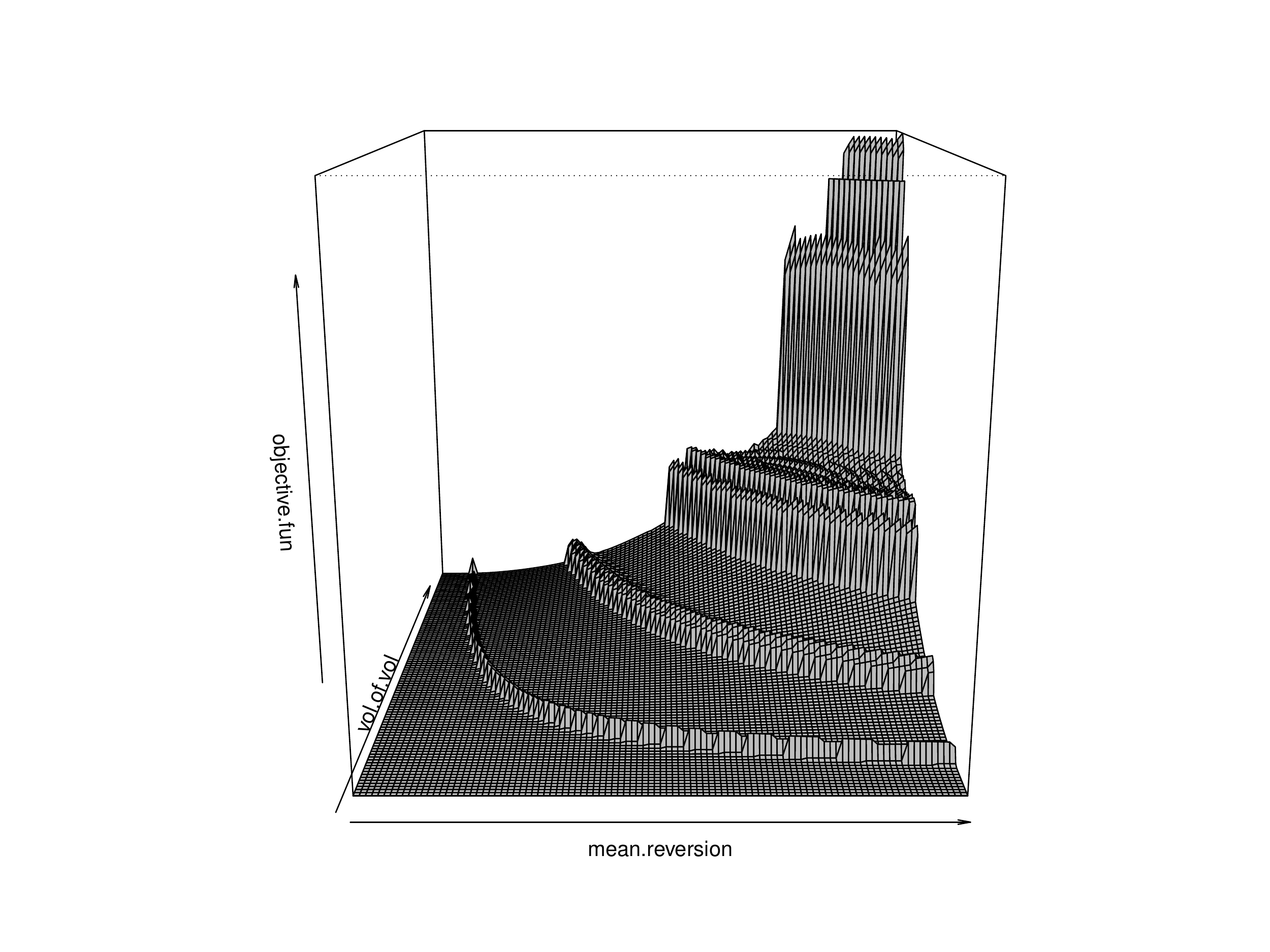}
\caption{Search space when varying volatility of volatility and rate of mean reversion}
\label{overflow1}
\end{figure}

\begin{figure}
\centering
\includegraphics[width=80mm]{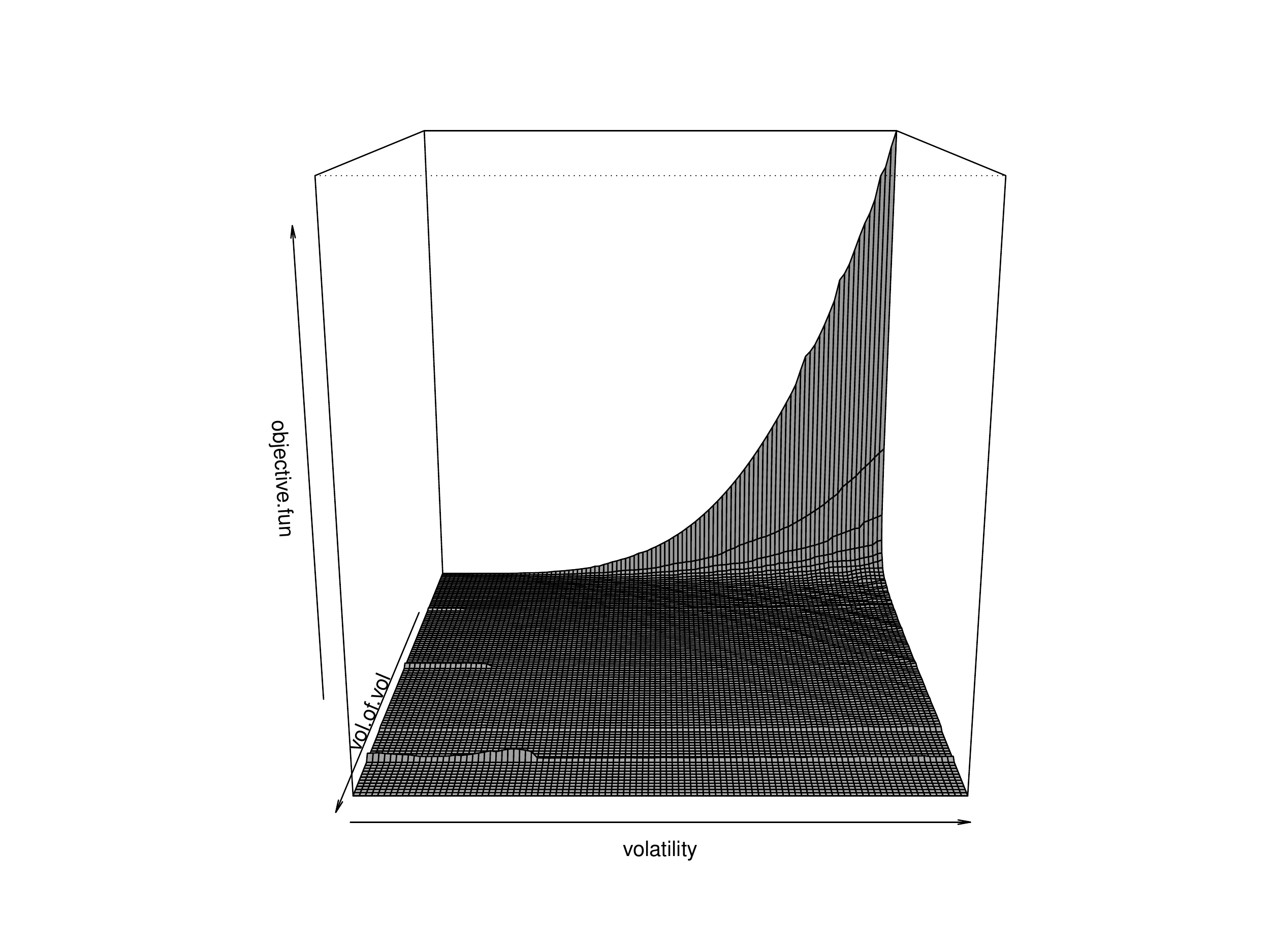}
\caption{Search space when varying volatility and volatility of volatility}
\label{overflow2}
\end{figure}

The search space for the optimization problem is bounded continuous and not convex, which means standard optimization methods (e.g. gradient methods based on derivates) are not applicable.
It is also characterized by a very high amount of local minima which complicates the task at hand even more. Figures \ref{overflow1} and \ref{overflow2} show the results of the objective function given in Equation (\ref{eq:objfunc}) when varying certain parameters.

As can be seen from the figures, the optimization method must find the perfect balance between exploration (traverse the whole search space relatively quickly) and exploitation (search the near-neighbour regions of the best found solutions so far) to avoid getting trapped in the many local minima.

In order for the algorithm to be a viable option, it has to fulfill the exploration-exploitation criteria as well as be numerically stable, meaning the error from the obtained solution and the actual one has to be minimal. Another important criterion is computational efficiency. The algorithm has to converge to the solution as quickly and as stable as possible.

The complexity of this task for the Heston pricing model for American options has so far prevented an effective and reliable method to arise. To this date, no algorithm has been proposed to solve this specific problem. In order to fulfill all of the above-mentioned criteria, a heuristic method called Cuckoo search will be used to calibrate the Heston option pricing model to a set of sample data.

\section{Cuckoo Search}
\label{sec:cuckoo}

The Cuckoo search algorithm is a naturally-inspired optimization algorithm developed by Yang and Deb \cite{Cuckoo}. It is a so-called meta-heuristic algorithm meaning that it makes only few assumptions about the problem at hand and can be used effectively when there is only incomplete information and limited computational capacity given. This makes it an ideal candidate for the optimization problem outlined in section \ref{sec:problem}.

The heuristic used in this paper draws its inspiration from the behaviour of cuckoos that lay their eggs in nests of other birds that they come across during their flights. This so-called brood parasitism is translated into a very efficient optimization method. The cuckoo search algorithm follows three simple rules:

\begin{enumerate}
\item Every cuckoo lays one egg in a random nest.
\item The best eggs are carried over to the next generation.
\item Cuckoo eggs are discovered by the host bird with a probability $p_a \in [0,1]$.
\end{enumerate}

Here, the nests represent the amount of different solutions at each iteration of the optimization process. The eggs represent the solutions to the objective function and are obtained from the flight paths of the cuckoos. At each iteration there is a probability that a new solution gets discarded and replaced by a random result. The pseudo-code for the standard Cuckoo Search algorithm is given in Table  \ref{tab:algo}.

\begin{table}
\textbf{begin}\\
\-\hspace{0.5cm} \textit{Objective function f(x), x = $(x_1, \ldots, x_d)^T$}\\
\-\hspace{0.5cm} \textit{Initialize population of n host nests $x_i$ (i = 1, 2, \ldots, n)}\\
\-\hspace{0.5cm} \textbf{while} \textit{stopping criterion is not met}\\
\-\hspace{1cm} \textit{Get a cuckoo randomly by L{\'e}vy flights}\\
\-\hspace{1cm} \textit{Evaluate its fitness $F_i$}\\
\-\hspace{1cm} \textit{Choose a nest j among n randomly}\\
\-\hspace{1cm} \textbf{if} \textit{($F_i > F_j$)}\\
\-\hspace{1.5cm} \textit{replace j by the new solution}\\
\-\hspace{1cm} \textbf{end}\\
\-\hspace{1cm} \textit{A fraction ($p_a$) of worse nests are abandoned and replaced by new ones}\\
\-\hspace{1cm} \textit{Rank solutions and find the current best}\\
\-\hspace{0.5cm} \textbf{end while}\\
\-\hspace{0.5cm} \textit{Postprocess results and visualization}\\
\textbf{end}
\caption{Cuckoo Search pseudo-code}
\label{tab:algo}
\end{table}

To model the flight paths that lead to new solutions, L{\'e}vy flights are used (for an introduction to L{\'e}vy flights see \cite{Levy}) There is a plethora of research available on how the flight behaviour of insects, animals and even humans show the same characteristics as L{\'e}vy Flights (see for example \cite{levy2}, \cite{levy3}).

A L{\'e}vy flight is essentially a random walk where the step lengths are drawn from a L{\'e}vy distribution, which is heavy-tailed. When the L{\'e}vy flight is defined in a dimension greater than $1$, the steps are isotropic and random,meaning they are uniform in all orientations.

\begin{figure}
\centering
\includegraphics[width=75mm]{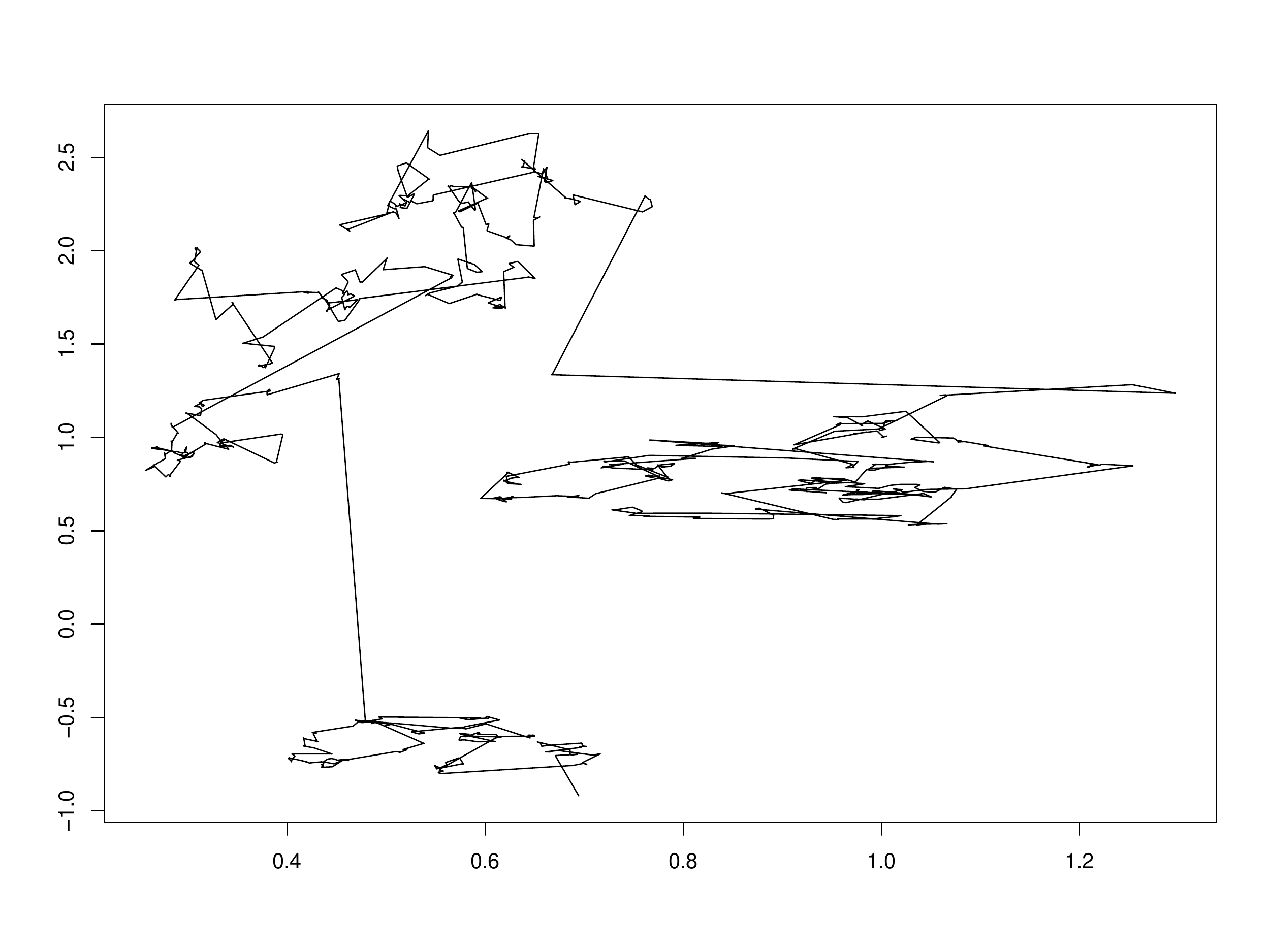}
\caption{L{\'e}vy flight in 2 dimensions with 1000 steps}
\label{fig:levyflight}
\end{figure}

The distinct feature of a L{\'e}vy flight is made visible in Figure \ref{fig:levyflight}. L{\'e}vy flights are characterized by large jumps followed by a series of small steps around the end of the jumps. This is exactly the behaviour needed for our optimization algorithm since it presents the perfect mix between exploration (the long jumps) and exploitation (the series of small steps). The long jumps also guarantee that the algorithm won't get trapped in the many local minima present in the search space.

A detailed explanation of the implementation of the Cuckoo search algorithm used in this paper is presented in the next section.

\section{Implementation}
\label{sec:implementation}

The implementation of the Heston pricing model itself will not be discussed here since any method that fulfills the criteria of robustness and numerical stability as well as fast computation time is suitable. For this to achieve, the pricer should be implemented in C++, Julia, Python or any similarly fast language.

The Cuckoo Search algorithm used in this paper has been implemented as an R \cite{R2014} package by the authors. Runtime tests using the package on standard optimization problems (Schwefel's Function, Rastrigins Function, Easoms Function) were solved in sufficient time for the Finance application.

In its essence, the implemented algorithm follows the steps laid out in the pseudo-code given in Table \ref{tab:algo}. It is modified in some parts to increase its effectiveness and convergence.

The implemented method requires a small number of parameters because of its heuristic nature. The necessary input includes the number of nests (starting points and different solutions at every iteration), the number of iterations (or the error tolerance in the objective function), upper and lower bounds for the parameters of the Heston pricing model as well as values $p_{max}$, $p_{min}$ and $p_{base} $ that are needed for the calculation of $p_{\alpha}$.

The part where the chosen implementation differs from the original algorithm is in the way $p_a$, the probability that the host bird discovers the cuckoos egg, is set. In the original paper, this is a fixed value. However Valian, Tavakoli, Mohanna and Haghi \cite{Improved} show some modifications that improve accuracy and convergence of the cuckoo search algorithm. One of these modifications is that $p_a$ is not fixed but a decreasing function between two set values, $p_{max}$ and $p_{min}$. The function looks as follows:

\begin{equation}
p_a (iter) = p_{max} - \frac{(p_{max} - p_{min})}{N} \times iter,
\end{equation}

where $N$ is the total number of iterations and $iter$ is the current iteration. This guarantees that the algorithm finds many new solutions in the beginning (the exploration is enhanced) and gradually moves on to find new solutions in the neighbourhood of points with an already good fitness (the exploitation is enhanced). Here we set the value for $p_{max}$  to 0.95 to have a lot of new solutions in the beginning of the optimization to further increase the exploration. The value for $p_{min}$ is set to 0.05 so that at the end of the optimization the algorithm searches almost exclusively around the best found positions.

The fitness function used to evaluate each solution in the search space calculates simply the sum of squared errors between the prices calculated by the Heston model and the observed (real) prices. It takes as inputs the parameters given by an egg in the Cuckoo Search algorithm and calculates the prices for all the asset value/strike price combinations.

\section{Numerical Results}
\label{sec:results}

To test and calibrate the model, an artificial dataset has been used to validate the solutions. The parameters used in creating the dataset have been chosen to be similar to the dataset in Natural Computing Vol 4, Chapter 2 \cite{Natural4}. The model has been tested varying the spot price $S_0$ as well as the strike price, K. In both occasions, the fixed price is set to 100 and the other price varies between 80 and 120 in steps of 2.

The risk-free interest rate has been set to $5\%$, however it is very easy to adapt the model to allow for a varying interest rate as well. The maturities $\tau$ for the options are $\frac{1}{12}$, $\frac{1}{4}$, $\frac{1}{2}$ and 1 year. This results in a total amount of 84 prices that are used to calibrate the model.

The parameter sets of the Heston pricing models are given in Table \ref{tab:parameters}.

\begin{table}
\begin{center}
\begin{tabular}{| c | c | c | c | c|}
  \hline                       
  $\sqrt{v_0}$ & 0.2 & 0.5 & 0.3 & 0.4\\
  \hline
  $\sigma$ & 0.1 & 0.1 & 0.25 & 0.25\\
  \hline
  $\kappa$ &  3 & 3 & 2 & 1\\
  \hline
  $\theta$ & 0.04 & 0.25 & 0.09 & 0.16\\
  \hline
  $\rho$ & -0.1 & -0.5 & -0.1 & -0.2\\
  \hline
	
\end{tabular}
\caption{Parameter sets for the Heston pricing model}
\end{center}
\label{tab:parameters}
\end{table}

Lower and upper bounds for all 5 parameters based on logic and empirical findings have been introduced to restrict the search space only to plausible solutions. The boundaries are as shown in Table \ref{tab:bounds}.

\begin{table}
\begin{center}
\begin{tabular}{| c | c | c | c | c | c|}
  \hline                       
  & $\sqrt{v_0}$ & $\sigma$ & $\kappa$ & $\theta$ & $\rho$\\
  \hline
  Lower Bound &  0.01 & 0.05 & 0.5 & 0.01 & -0.8\\
  \hline
  Upper Bound & 0.1 & 0.3 & 4 & 0.1 & 0.1\\
  \hline
	
\end{tabular}
\caption{Lower and upper bounds for the parameters.}
\label{tab:bounds}
\end{center}
\end{table}

For the correlation parameter, $\rho$, that captures the correlation between volatility and the stock price, the range has been set to be mostly negative according to several papers focused on this topic, e.g. Nandi \cite{Correlation}.

For the choice of the number of nests used, multiple values have been tested. The algorithm has proved itself being very stable to changes in the number of nests, meaning values between 20 and 50 have all shown fast convergence and high accuracy. 

\begin{figure}
\centering
\includegraphics[width=75mm]{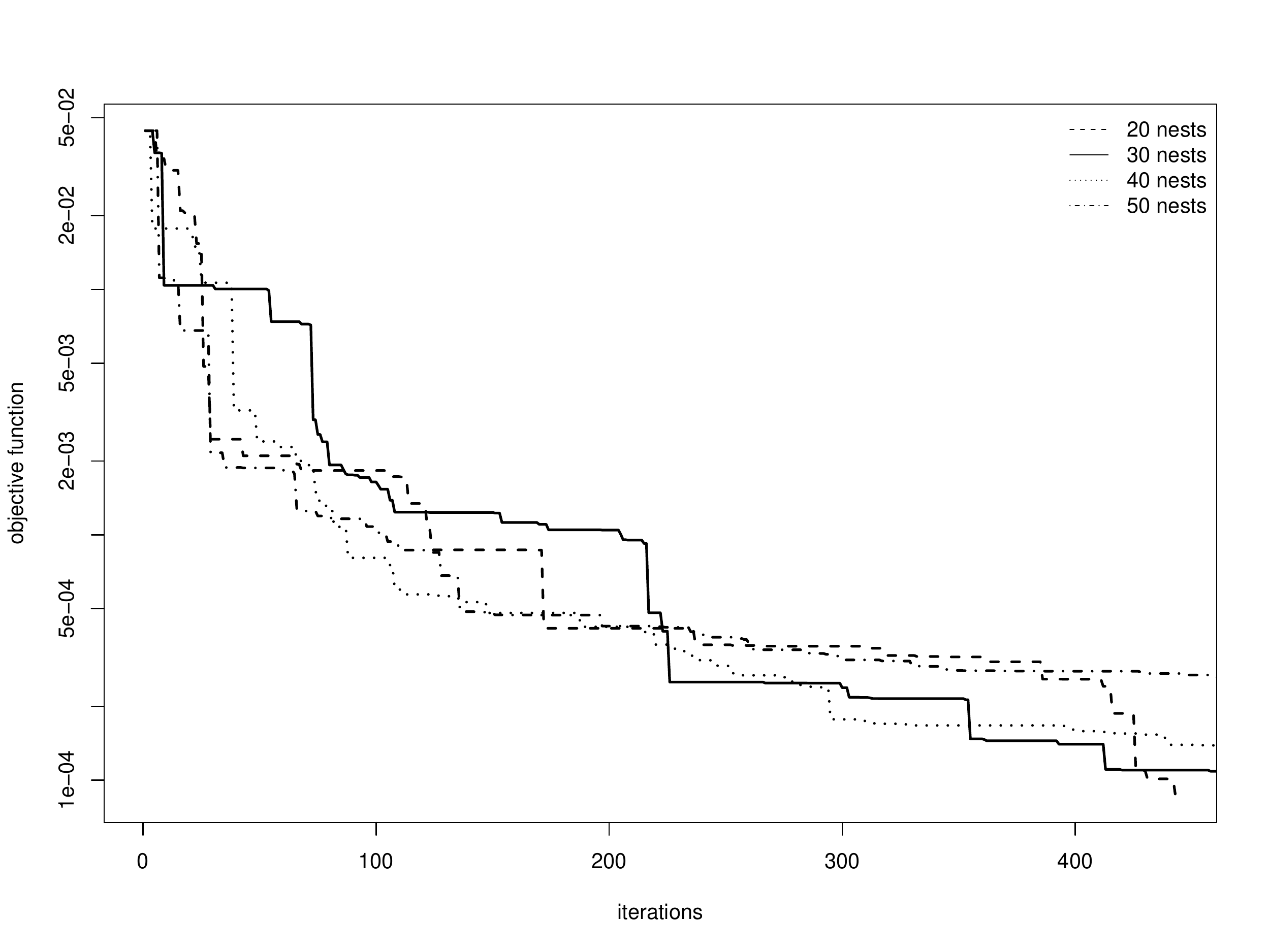}
\caption{Convergence comparison for different nest sizes.}
\label{fig:overflow}
\end{figure}

From Figure \ref{fig:overflow} it is visible that for all choices of nests, convergence of the objective function is given and is relatively fast. This means there is no need to fit this parameter for the model and its selection can be made entirely to increase computational efficiency. The selection made for this paper is a nest count of 20.

\begin{figure}
\centering
\includegraphics[width=75mm]{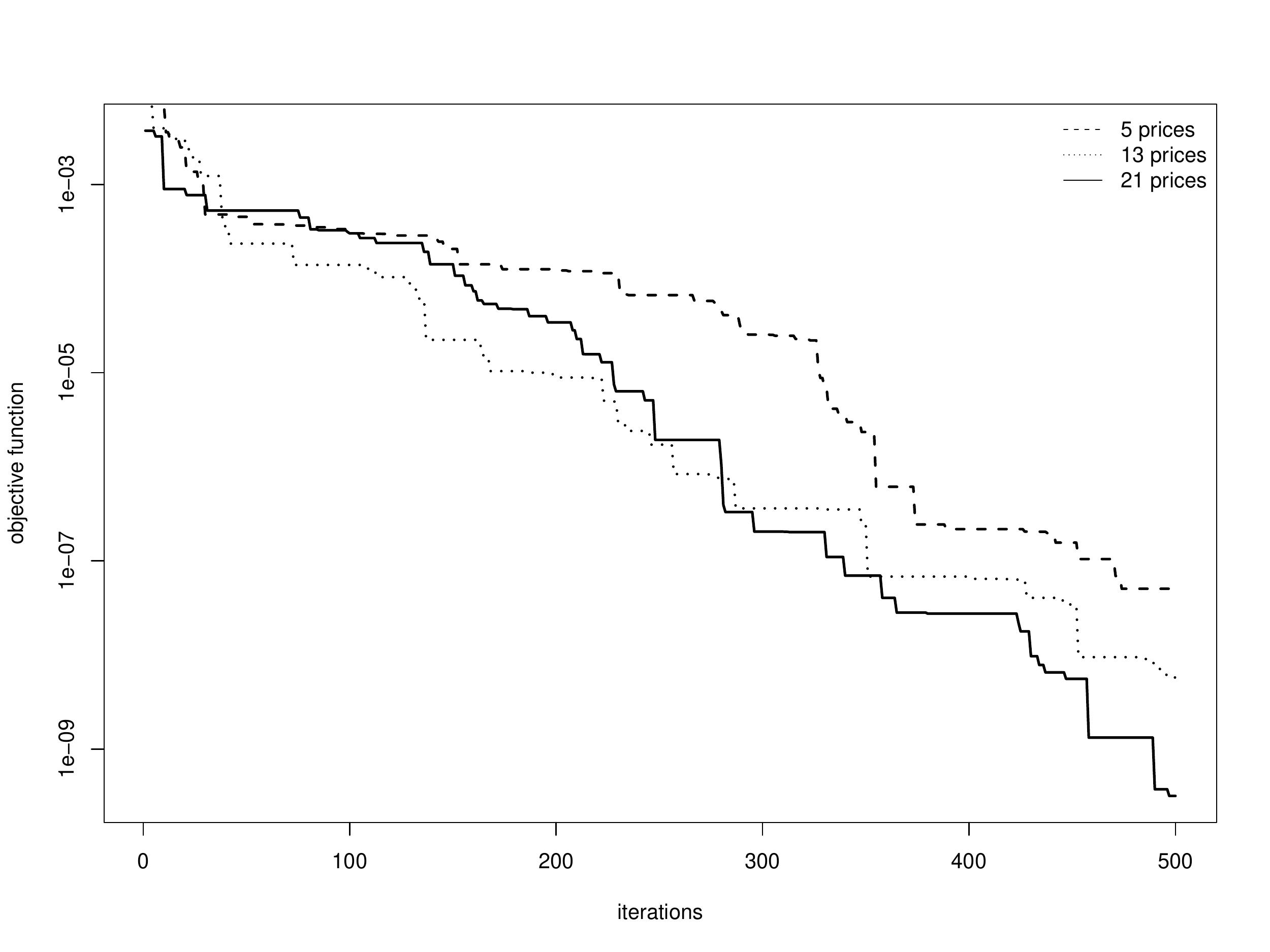}
\caption{Convergence comparison for different number of prices}
\label{fig:priceconv}
\end{figure}

Furthermore tests on how many prices the algorithm needs as input for calibration have been conducted. Figure \ref{fig:priceconv} shows the convergence on a log-scale for inputs of 5,13 and 21 prices. It is clearly visible that the algorithm needs fewer iterations if the number of prices increases. There is, however, an increase in computation time for more prices.

\begin{figure}
\centering
\includegraphics[width=100mm]{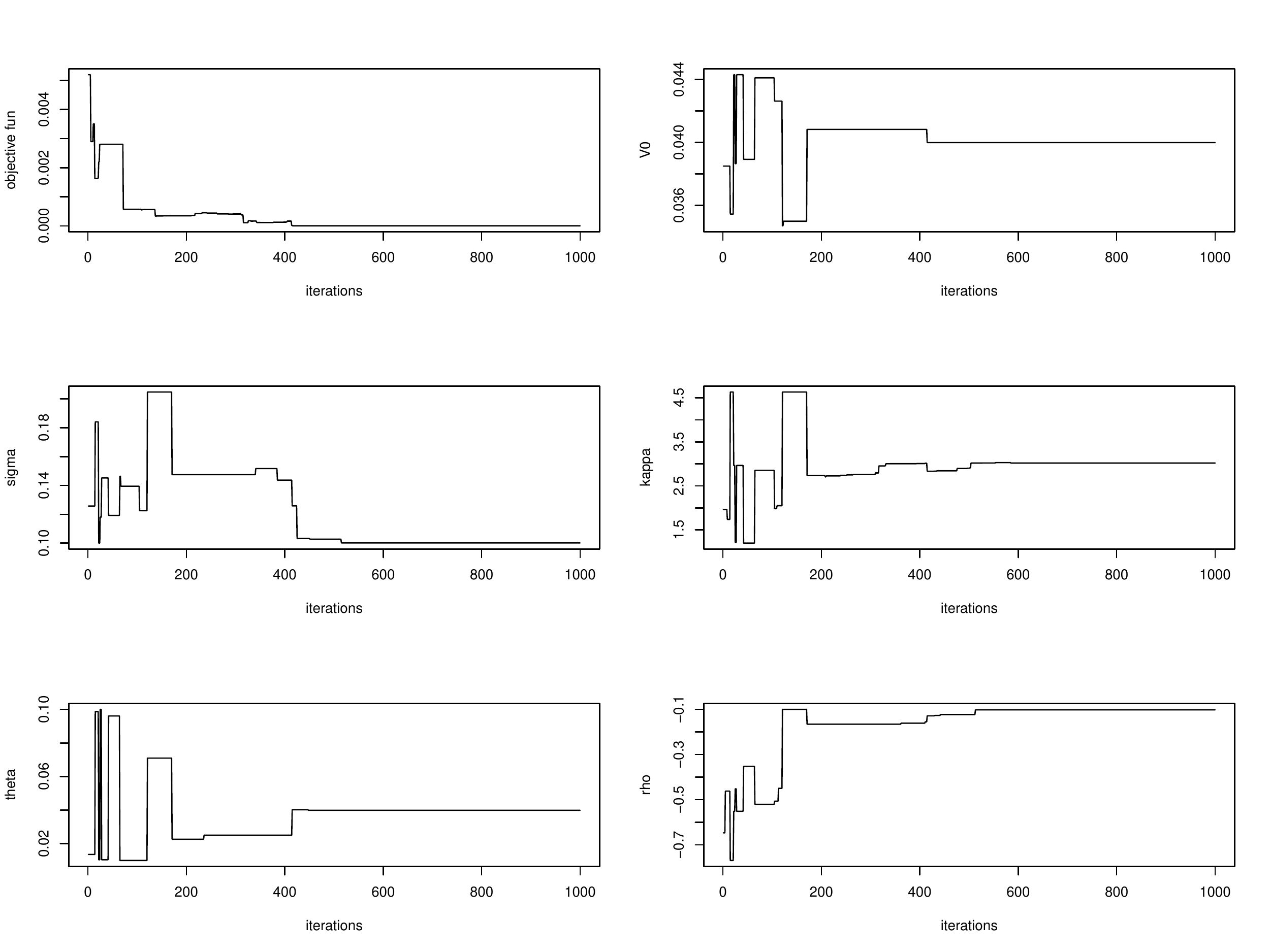}
\caption{Convergence of objective function and parameters.}
\label{fig:allconv}
\end{figure}

The graphs shown in Figure \ref{fig:allconv} show the convergence of the objective function as well as of all 5 parameters calibrated by the Cuckoo Search algorithm. It can be seen, that there is a fast and stable convergence not only of the objective function but also of the parameters. The values of the objective function and the parameters after conducting the optimization process for each of the parameter sets are given in Table \ref{tab:objective}.

\begin{table}
\begin{center}
\begin{tabular}{|c|c|c|c|c|c|c|}
\hline
& Objective function & $\sqrt{v_0}$ & $\sigma$ & $\kappa$ & $\theta$ & $\rho$ \\
\hline                       
CS & 0.000001 & 0.2 & 0.1 & 2.999 & 0.04 & -0.09998 \\
True & - & 0.2 & 0.1 & 3 & 0.04 & -0.1 \\
\hline
\hline
CS & 0.000001 & 0.49999 & 0.09999 & 3.0012 & 0.25 & -0.5 \\
True & - & 0.5 & 0.1 & 3 & 0.025 & -0.5 \\
\hline
\hline
CS & 0.000001 & 0.3 & 0.250098 & 2 & 0.089978 & -0.1 \\
True & - & 0.3 & 0.25 & 2 & 0.09 & -0.1 \\
\hline
\hline
CS & 0.000001 & 0.39998 & 0.25 & 0.99875 & 0.16 & -0.200125 \\
True & - & 0.4 & 0.25 & 1 & 0.16 & -0.2 \\
\hline
	
\end{tabular}
\caption{Objective function and parameter values after optimization.}
\label{tab:objective}
\end{center}
\end{table}

\section{Conclusion}
\label{sec:conclusion}

In this paper an improved Cuckoo Search Algorithm has been developed to allow for an efficient and robust calibration of the Heston option pricing model for American options. Calibration of stochastic volatility models like the Heston is significantly harder than classical option pricing models as more parameters have to be estimated. The difficult task of calibrating one of these models to American Put options data has been solved using an implementation with the statistical programming langauge R. The numerical results are very promising such that future research will focus on applying the algorithm to real-world Options data.

\end{document}